\documentclass[10pt,twocolumn,letterpaper]{article}

\usepackage{cvpr}
\usepackage{times}
\usepackage{epsfig}
\usepackage{graphicx}
\usepackage{amsmath}
\usepackage{amssymb}
\usepackage{subcaption}
\usepackage{caption}
\usepackage{multirow}
\usepackage{booktabs}
\usepackage{afterpage}

\cvprfinalcopy 

\def\cvprPaperID{10553} 

\ifcvprfinal\fi
\begin{document}

\title{EDAS: Efficient and Differentiable Architecture Search}

\author{Hyeong Gwon Hong,$^{2}$ Pyunghwan Ahn$^{1}$, Junmo Kim$^{1}$\\
$^{1}$School of Electrical Engineering, KAIST\\
$^{2}$Graduate School of AI, KAIST\\
{\tt\small \{honggudrnjs,dksvudghks,junmo.kim\}@kaist.ac.kr}
}

\maketitle

\begin{abstract}
  Transferrable neural architecture search can be viewed as a binary optimization problem where a single optimal path should be selected among candidate paths in each edge within the repeated cell block of the directed acyclic graph form. Recently, the field of differentiable architecture search attempts to relax the search problem continuously using a one-shot network that combines all the candidate paths in search space. However, when the one-shot network is pruned to the model in the discrete architecture space by the derivation algorithm, performance is significantly degraded to an almost random estimator. To reduce the quantization error from the heavy use of relaxation, we only sample a single edge to relax the corresponding variable and clamp variables in the other edges to zero or one. By this method, there is no performance drop after pruning the one-shot network by derivation algorithm, due to the preservation of the discrete nature of optimization variables during the search. Furthermore, the minimization of relaxation degree allows searching in a deeper network to discover better performance with remarkable search cost reduction (0.125 GPU days) compared to previous methods. By adding several regularization methods that help explore within the search space, we could obtain the network with notable performances on CIFAR-10, CIFAR-100, and ImageNet.
\end{abstract}

\begin{figure}[ht!]
\centering
\includegraphics[width=1.0\linewidth, page=1]{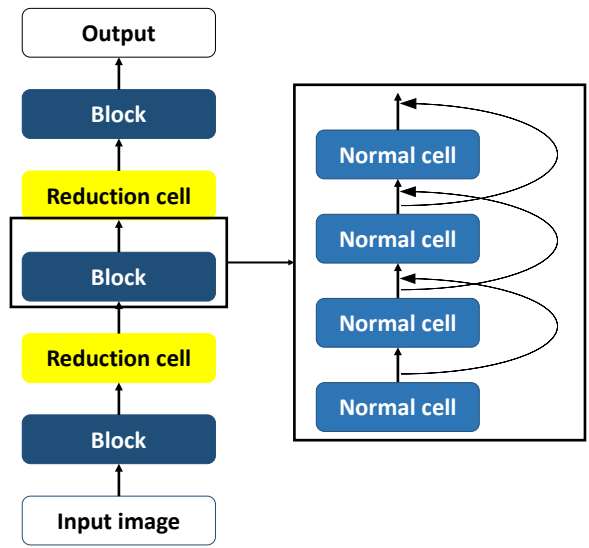}
\caption{Neural architecture search problems are usually downscaled as a cell search problems, and the overall network consists of several blocks composed of a certain number of normal cells and reduction cells located between two adjacent blocks.
}
\label{fig:network_cell}
\end{figure}

\section{Introduction} \label{sec1}

Over the last several years, convolutional neural networks (CNNs) have shown remarkable performances in many computer vision tasks, including image classification~\cite{krizhevsky2012imagenet,simonyan2014very,szegedy2015going,he2016deep,huang2017densely}, object detection~\cite{girshick2014rich,ren2015faster,redmon2016you}, and semantic segmentation~\cite{long2015fully,noh2015learning,chen2017deeplab}. Researchers have invested significant effort into searching for proper architectures for each task to achieve better performance. However, because each new task requires a new architecture, applying CNNs to computer vision tasks remains computationally expensive. Thus, recent studies have attempted to automate the process of searching for better architectures.

Recently, most Neural Architecture Search (NAS) methods perform search on a cell structure (Figure~\ref{fig:network_cell}), which is a building block of networks, since searching for the whole network is computationally expensive. The cell structures are represented as graphs (Figure~\ref{fig:cell_structure}), in which nodes represent feature maps in neural networks and edges represent operations such as convolution, pooling, etc. Even though performing search on cells requires less computation, state-of-the-art methods based on reinforcement learning~\cite{zoph2018learning} or evolutionary algorithm~\cite{real2019regularized} still run at the cost of a large computational burden. Thus, they are not considered appropriate for new tasks or datasets. In the spirit of developing an efficient architecture search, DARTS (Differentiable ARchiTecture Search)~\cite{liu2018darts} suggested to train \textit{one-shot network} with special parameters that will be used in selecting architecture. Through simplification of the search algorithm, the running time of DARTS is three orders of magnitude shorter than NASNet or AmoebaNet while producing comparable results.

However, operation weights learned in the search stage of DARTS cannot be directly used for evaluation because the one-shot architecture is pruned to only select optimal path in the evaluation stage, discarding a large number of paths. The overparameterization of the one-shot network results in significantly different network behavior during search and evaluation. We verify this through evaluation of derived models in the middle of search stage. This gap between architectures is the reason that DARTS needs re-training of the derived network from scratch.

We propose a differentiable one-shot architecture search method in which only a single edge inside the cell is modified per iteration. At every iteration, our method samples one edge from the cell for execution of all candidate operations, while all other edges used to perform only one operation. For each edge that performs one operation, we select the operation with the highest architecture parameter at the time. Because most edges perform only one operation at each iteration, the network architecture used in our method is very similar to those used for evaluation.

Our method can be thought of as performing DARTS for a single edge per iteration, which has advantages over the original implementation of DARTS. First, our method does not suffer from the \textit{depth gap}, suggested by \cite{chen2019progressive}. Because we only need to calculate one operation of each edge, memory issues are alleviated, expanding the applicability of our algorithm to deeper network candidates. Second, the number of operations is reduced via single edge relaxation, reducing the search cost by a significant amount. Finally, because architectures are much more similar to those in the target search space, our method is more interpretable and shows better performance than DARTS. Through extensive experiments, we verify that our method shows comparable results with state-of-the-art search algorithms in 0.125 GPU days on several benchmark datasets.

\section{Related Works} \label{sec2}

Since the innovation of CNNs, led by AlexNet~\cite{krizhevsky2012imagenet}, some experts have tried designing architectures manually. Following \cite{krizhevsky2012imagenet}, architectures with smaller filter sizes~\cite{simonyan2014very}, multi-path structures~\cite{szegedy2015going}, or extremely deep networks~\cite{he2016deep} have been suggested. In more recent years, some other attempts have been made, such as widening the network instead of increasing depth~\cite{zagoruyko2016wide}, or densely connecting layers~\cite{huang2017densely} to ease the flow of the gradient.

Another family of studies has suggested NAS methods, which perform automated architecture search for specific tasks. There are two main streams in the development of NAS methods, including reinforcement learning-based methods~\cite{zoph2018learning,pham2018efficient} and evolution-based methods~\cite{real2019regularized}. In NASNet~\cite{zoph2018learning}, authors suggest sampling several architectures and training them to a certain extent to estimate performance, and then to decide how the architecture should be changed in order to maximize performance improvement. Because the process of training many architectures is computationally heavy, \cite{pham2018efficient} has proposed to share weights among the architectures to reduce training time. Some other works such as \cite{real2019regularized} shares sampling and evaluation process of NASNet, but use evolutionary algorithms as meta-controller.

More recently, DARTS~\cite{liu2018darts} was proposed as an efficient and effective algorithm, taking only 4 GPU days of search cost and showing results comparable to NASNet and AmoebaNet. DARTS is a one-shot architecture search method in which all candidate paths are included in a single network, which was originally suggested in \cite{bender2018understanding}. Operation parameters are trained along with architecture parameters with separate data. DARTS is extremely efficient compared to \cite{zoph2018learning} and \cite{real2019regularized}, as all operations are trained at once. Following works make various modifications based on DARTS to improve search ability.

SNAS~\cite{xie2018snas} argued that because the output of an edge in DARTS is always a weighted sum of several operations, it differs from the information flow expected at the evaluation stage. This problem is verified by dramatic performance drop after applying derivation algorithm in DARTS, which implies that network parameters are not properly trained. In \cite{xie2018snas}, stochastic NAS is suggested to alleviate this gap by applying gumbel softmax to the architecture parameters, for the network to behave as though only one operation has been applied to each input feature.

ProxylessNAS~\cite{cai2018proxylessnas} suggested to explicitly sample a single path from the one-shot network using a binary mask. This bridges the gap of cell structures in the search and evaluation stages completely. Also, it allows searching on the networks of the same depth on the same dataset as the evaluation stage, because only one operation needs to be calculated, which greatly alleviates memory issues in one-shot architecture search. In \cite{cai2018proxylessnas}, they also proposed to make latency limitation into differentiable loss function, so that hardware constraints can be considered in training. They achieved great performance improvement over state-of-the-art in CIFAR-10 and ImageNet.

\section{Method} \label{sec3}

\begin{figure}[ht!]
\centering
\includegraphics[width=0.7\linewidth,page=1]{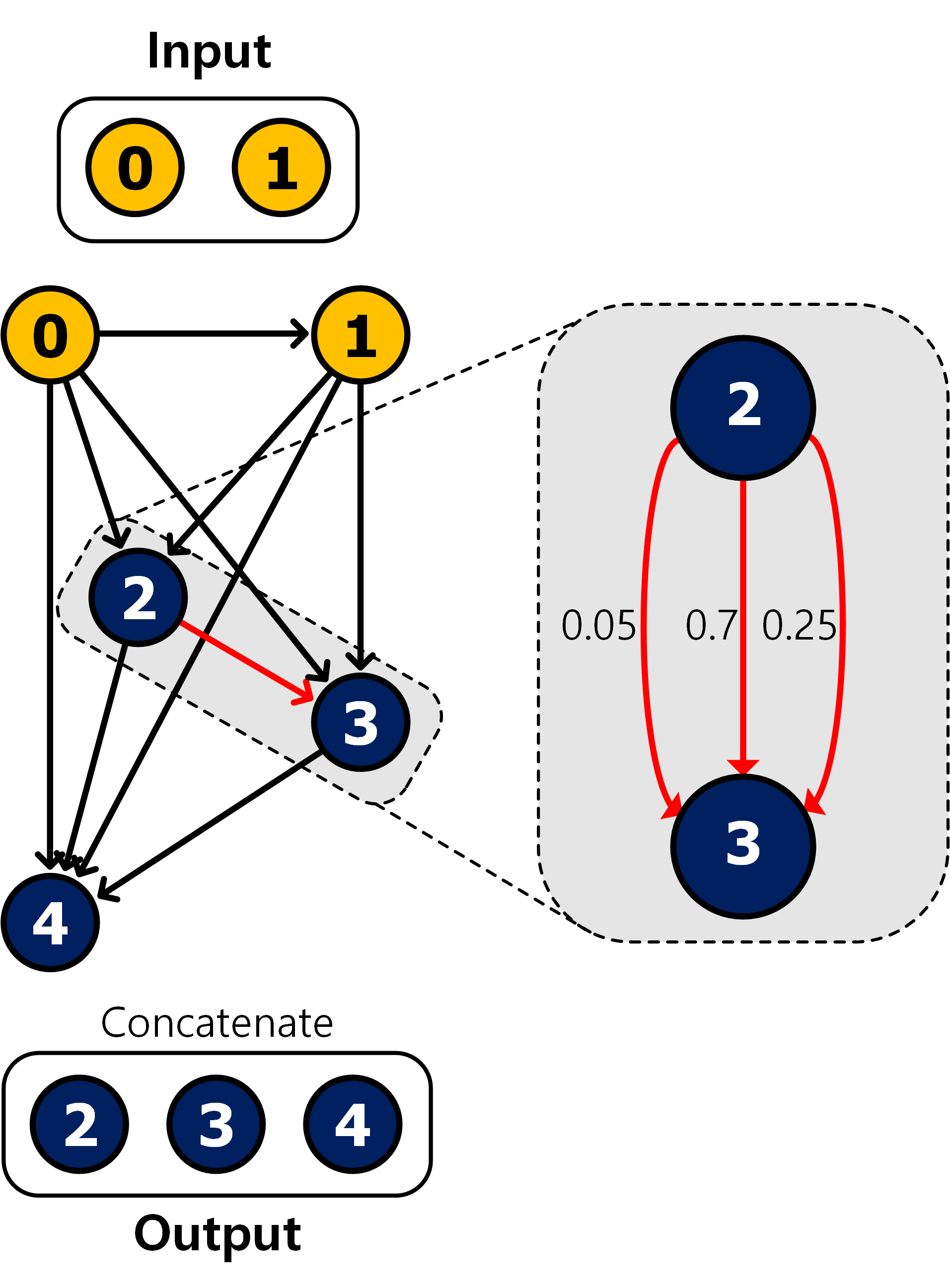}
\caption{Description for parameter training of our method. Basically, each edge (arrow) contains all the candidate operations in the search space along with architecture parameters ($\alpha$) that are learned to represent the superiority of each operation at the edge. In our method, only a randomly sampled single edge (red) updates architecture parameters and all the other edges (black) are forwarded with the operation that corresponds to the maximum architecture parameter. Our method forwards the green edge using continuous relaxation. In particular, all the candidate operations are mixed proportionally with as the softmax of architecture parameters. Our first method simply updates both architecture parameters and operation weights this way.
}
\label{fig:cell_structure}
\end{figure}

\subsection{Preliminary: One-shot architecture search} \label{sec3.1}

Our method is heavily related to one-shot architecture search~\cite{bender2018understanding} and a follow-up work DARTS~\cite{liu2018darts}, which are efficient search algorithms. In one-shot methods, all candidate operations for each edge are included in a single overparameterized network, and they search for the optimal architecture by training operation parameters and choose the one with the highest estimated performance. In DARTS, special parameters named as architecture parameters are also trained through gradient descent along with operation parameters. In this section, we explain how one-shot architecture search works and define notations that will be used in the following sections, based on DARTS.

In recent NAS methods, the network is represented as repeated cell structures which can be stacked as many times as necessary. Following recent well-performing CNN architectures~\cite{he2016deep,huang2017densely}, the network is divided into stages, each of which consists of the same cell structures (normal cell) repeated, and between those stages, reduction cells with different structure are inserted to downsample intermediate feature maps. A cell is represented by a directed acyclic graph (DAG) with $N_n$ nodes, typically using 7 for $N_n$, which consists of two input, one output, and four intermediate nodes. In \ref{fig:cell_structure}, we show simplified version of cell structure with only 2 intermediate nodes. In the DAG, each edge connects a pair of nodes $(i,j)$, where $0 \leq i < j \leq N_n-1$, and operates on the output of node $i$ to generate features for node $j$.

In DARTS, an edge calculates output as a weighted sum of all candidate operations:
\begin{equation}
f_j(x_i) = \Sigma_{o \in O} \frac{\exp{\alpha_o^{(i,j)}}}{\Sigma_{o' \in O} \exp{\alpha_{o'}^{(i,j)}}} * o(x_i)
\end{equation}
where $x_i$ is the output of the $i$-th node, $o$ is one of the candidate operations $O$, and $\alpha$ is the architecture parameter which will be used to select architecture after search is finished. For the candidate operation set $O$, we follow the set of $N_o = 7$ operations used in DARTS\cite{liu2018darts}. Each intermediate node is calculated as a sum of operations in all the edges between itself and preceding nodes. The final output of a cell is a concatenation of all intermediate nodes. This structure is illustrated in Figure~\ref{fig:cell_structure}.

DARTS is entirely differentiable because it applies the softmax function to architecture parameters ($\alpha$), thus relaxing the discrete search problem into a continuous one. This allows all parameters in the network to be trained by gradient descent. Bi-level optimization is utilized to train operation parameters and architecture parameters, using training data and separate validation data, respectively. After the search process, the cell structure is pruned with architecture parameters, leaving only one operation with the highest $\alpha$ at each edge. Then, the final network for evaluation is constructed by repeating the cell structure with a single operation per edge, and then trained from scratch.

\subsection{Our Method} \label{sec3.2}

\begin{figure}[ht!]
\centering
\includegraphics[width=1.0\linewidth,page=1]{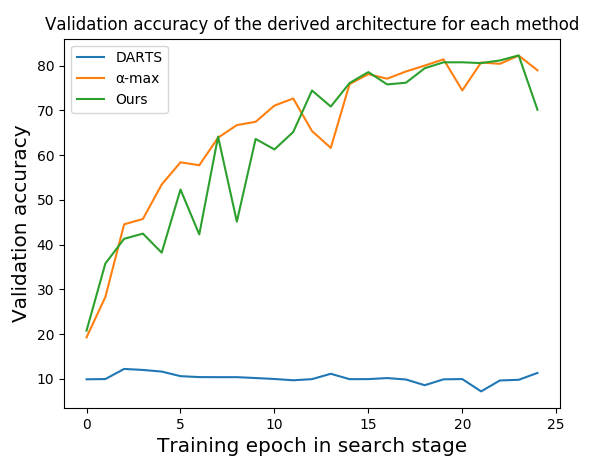}
\caption{Validation accuracy of the derived model for each iteration during \textit{search stage} for several methods: \textbf{DARTS}, \textbf{$\mathbf{\alpha}$-max}, and our method. The derivation algorithm based on architecture parameters is the same for all the methods. 
}
\label{fig:val_acc}
\end{figure}

In previous differentiable methods, all candidate operations are computed for every edge and then weighted sum of their results are stored as the output of the edge. However, only a single operation is applied at each edge in the evaluation stage. Thus, features are not propagated through the same path during search and evaluation. This results in a significant difference in network behavior, which is experimentally shown as the gap of performance between the networks before and after architecture derivation~\cite{xie2018snas}. We also verify that the derived model in DARTS always shows poor validation accuracy throughout the training process, as shown in Figure~\ref{fig:val_acc}. This supports the fact that individual operations are not meaningful during search in DARTS. Also, training operations with inaccurate gradients would disturb fast training of the target network, and thus result in increased search cost.

Based on these observations, we propose to sample one operation per edge. For each edge, the operation with the maximum architecture parameter ($\alpha$) is sampled so that the network exactly represents the derived model at the moment. Then, operation parameters are trained in this network, which prevents the performance gap that would otherwise occur in training between the one-shot network and the derived architecture. However, this limits the advantages of one-shot methods, which normally have simultaneous training of all operation parameters. If only one operation is trained at each iteration, the architecture parameter of that operation is more likely to be increased and thus the other operations have limited opportunity of being trained. We verify this by tracking architecture parameters over the training procedure and confirm that top ranked operation rarely changes (Figure~\ref{fig:alpha}).

As a compromise between both methods, we suggest performing one-shot architecture search in only one edge per iteration. For example, as shown in Figure~\ref{fig:cell_structure}, we choose one edge (red arrow) and train all parameters in that edge. Meanwhile, the other edges (black arrow) simply pass the information forward through only the maximum-$\alpha$ operation. In this way, we can perform the search on architectures that are much closer to those used in evaluation stage. This procedure can be represented by the following equation:
\begin{equation}
f_j(x_i) = \begin{cases}
\Sigma_{o \in O} \frac{\exp{\alpha_o^{(i,j)}}}{\Sigma_{o' \in O} \exp{\alpha_{o'}^{(i,j)}}} * o(x_i) &\text{$i=k,  j=l$}\\
o_m(x_i) &\text{otherwise}
\end{cases}
\end{equation}
where $k$ and $l$ are two nodes connected by the selected edge, and $o_m$ is selected as the operation with the maximum architecture parameter $\alpha$.

This method has the advantages of both one-shot architecture search and sampling-based architecture search. Our proposed method does not suffer from the performance gap between the one-shot model and the derived model. Furthermore, non-maximum $\alpha$ operations are trained along with the maximum $\alpha$ operation, so that the other edges can be constantly affected by those operations and get prepared for the change of architectures. Thus, the change of top rank operation happens more often under our method, as shown in Figure~\ref{fig:alpha}.

Additionally, we alternately inverse architecture parameters at the beginning of training to prevent dominant training of operations with larger initial $\alpha$. Methods based on sampling, either in all edges or most of them, suffer from limitation of exploring operations with lower architecture parameter at the initial state. This is shown in \ref{fig:alpha} as changing values of architecture parameters over epoch. The red!! line in \ref{fig:alpha} shows a case that suffers from limited training of operation parameters with non-maximum $\alpha$, which results in rare change of first rank operation. Based on this observation, we propose to periodically use $-\alpha$ as architecture parameters at the beginning of search stage, for better exploration over all parameters. When using $-\alpha$, the edge $(i,j)$ is computed as follows:
\begin{equation}
f_j(x_i) = \Sigma_{o \in O} \frac{\exp{(-\alpha_o^{(i,j)})}}{\Sigma_{o' \in O} \exp{(-\alpha_{o'}^{(i,j)})}} * o(x_i)
\end{equation}
With this modification to the algorithm, second rank operations are more capable of turning into the first, as shown in \ref{fig:alpha}.

Our method is motivated by the coordinate descent algorithm used in the discrete image hashing problem. Learning-based discrete image hashing is effective for efficient storage or retrieval of high-dimensional image data. The typical difficulty in this field is to preserve the discrete nature of a hash code variable while it is relaxed to a real variable during training to improve computational efficiency. One of the solutions to this problem is cyclic coordinate descent where only a single variable is relaxed to be optimized, while the other remaining variables are clamped to zero or one for a binary case. In this way, search accuracy loss from heavy relaxation can be avoided, and more efficient optimization can be achieved. Thus, because architecture search can be considered as a binary variable optimization like the hashing problem, we suggest a coordinate descent algorithm to solve the problem of heavy relaxation in DARTS.

\subsection{Comparison to other methods} \label{sec3.3}

Since one-shot architecture search~\cite{bender2018understanding} was proposed, many efficient and differentiable neural architecture search algorithms have been suggested, and some of them are related to our method. Thus, in this section, we compare our method to previous works that are similar to ours.

Our method is heavily motivated by DARTS~\cite{liu2018darts}. In DARTS, all candidate operations in the search space are calculated in all edges, which causes huge memory consumption for multiple feature maps in every node. We significantly alleviate this problem by computing only one operation in most edges while retaining the effectiveness of one-shot search methods. Because we sample a single operation, our method does not suffer from the huge performance gap between networks at architecture derivation, as mentioned in \ref{sec3.2}.

ENAS~\cite{pham2018efficient} proposed weight sharing scheme to minimize redundant procedure in exhausting training of different architectures. Our method is similar in that it samples different network architectures at each iteration, but weights are shared over architectures and stored in a single overparameterized network. Our method also leverages the architecture derivation process in DARTS, which utilizes learned architecture parameters, while ENAS trains a controller with policy gradient to select architectures.

ProxylessNAS~\cite{cai2018proxylessnas} is another similar method to ours in that it samples a single path in the network with binary variables and trains only the weights in the sampled path. ProxylessNAS aims to search architecture without any proxy tasks, such as training on a smaller dataset or smaller network. Our method has a different goal, as we still perform the search on a smaller dataset if the target dataset is of a large scale. However, we do not train on a proxy network with a smaller number of cells, as our method does not suffer from memory issues caused by storing all operation results in the one-shot network. Furthermore, when training on the target task, our method does not suffer from the depth gap~\cite{chen2019progressive} because we do not use a proxy network for the search stage.

\section{Experiments} \label{sec4}

Adopting the pipeline of DARTS, the experimental demonstration of our method consists of three stages: \textit{i) search stage}, \textit{ii) evaluation stage}, and \textit{iii) transfer stage}.

In the \textit{search stage}, we search for the optimal cell using our method. The main difference of our method relative to DARTS is that our method can search for the optimal cell in a deeper network using a single GPU, removing the need for layer extension in the evaluation stage, which DARTS requires to obtain better performance. Furthermore, we apply the warm-up period in the first few epochs of the search stage to obtain better results by reversing the sign of architecture parameters at every iteration to give a fair learning opportunity to each candidate operation. We demonstrate the exploration power of this technique by analyzing the architecture parameter learning during the search stage.

In the \textit{evaluation stage}, we retrain the network based on the cell structure found by our search algorithm from scratch and compare its accuracy to that of other methods. Then, the network is evaluated on a large-scale dataset for a transferability test in the \textit{transfer stage}. 

\subsection{Search stage} \label{sec4.1}

\noindent \textbf{Search space} The search space defines the architecture search problem, and it is determined by the set of candidate operations. As in recent works such as DARTS or PDARTS, we include the following candidate operations: 3x3 and 5x5 separable convolutions, 3x3 and 5x5 dilated separable convolutions, 3x3 max pooling, 3x3 average pooling, skip connection, and zero. The search algorithm can then find the optimal operation for each edge among candidate operations.

\noindent \textbf{One-shot network depth} In DARTS, a one-shot network composed of 8 cells (2 cells for each of the 3 stages and 2 reduction cells for downsampling between stages) is trained by optimizing operation weights and architecture parameters in an iterative manner (i.e., bilevel optimization). In our method, a stochastic single-edge modification allows the use of a deeper one-shot network composed of 20 cells (6 cells for each of the 3 stages with 2 reduction cells for downsampling between stages) despite the 12GB memory constraint of a single GPU memory constraint.

\noindent \textbf{Datasets} We searched for the optimal cell on the CIFAR-10 and also CIFAR-100, as in PDARTS, which is the first work that reported search performance on CIFAR-100. 

\noindent \textbf{Training settings} The one-shot network was trained for 25 epochs with a batch size of 64 using bilevel optimization adopted from DARTS, but only training architecture parameters for a single edge that was sampled randomly for each iteration. Operation parameters were trained by the Stochastic Gradient Descent (SGD) method with a momentum 0.9, weight decay rate 3e-4, and learning rate scheduled by cosine annealing learning rate scheduler. In particular, initial learning rate is 0.025 minimum learning rate is 0.001. For architecture parameters, they are trained by Adam optimizer with learning rate 0.025, betas (0.5, 0.999), and \textbf{zero weight decay rate}. The weight decay regularization is removed to prevent the architecture parameters for other unselected edges from being updated by weight regularize and focus only on modification of architecture parameters for the selected edge in our method. However, in the case of initial warm-up setting, weight decay rate is nonzero and set to 1e-3 as in DARTS to inhibit the architecture parameters from quickly selecting the optimal cell. In addition, the warm-up trick (inversing architecture parameters) is applied for every iteration in the first 10 epochs of \textit{training stage} to guarantee fair learning of candidate operations. 

   \begin{figure*}[htbp] 
     \centering
     \begin{subfigure}[b]{0.3\textwidth}
       \centering
       \includegraphics[width=\textwidth]{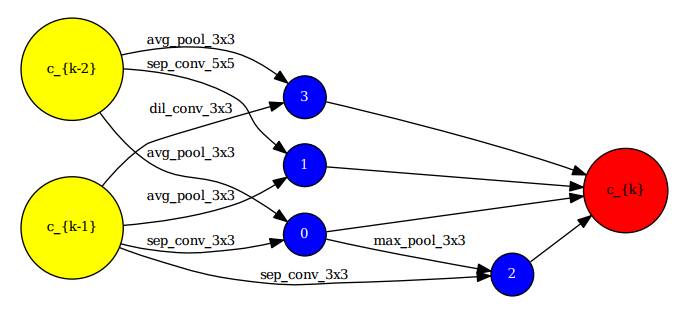}  
       \caption{Epoch 0}
       \label{fig:sub-first}
     \end{subfigure}
     \begin{subfigure}[b]{0.3\textwidth}
       \centering
       \includegraphics[width=\textwidth]{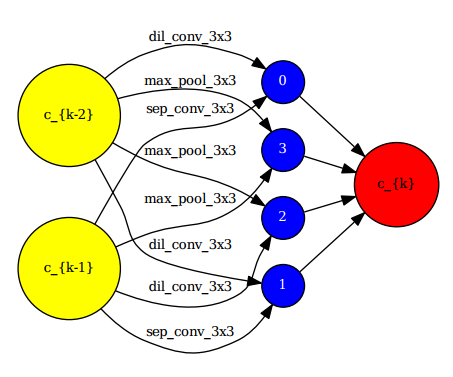}  
       \caption{Epoch 20}
       \label{fig:sub-second}
     \end{subfigure}
     \begin{subfigure}[b]{0.3\textwidth}
       \centering
       \includegraphics[width=\textwidth]{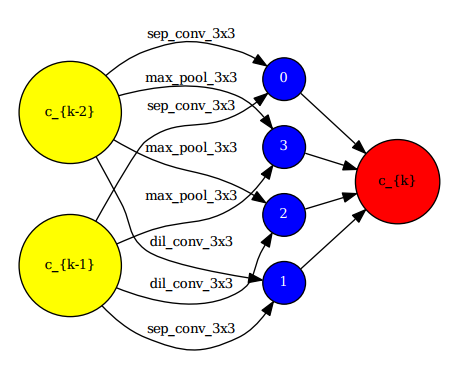}
       \caption{Epoch 40}
       \label{fig:sub-second}
     \end{subfigure}
     \begin{subfigure}[b]{0.3\textwidth}
       \centering
       \includegraphics[width=\textwidth]{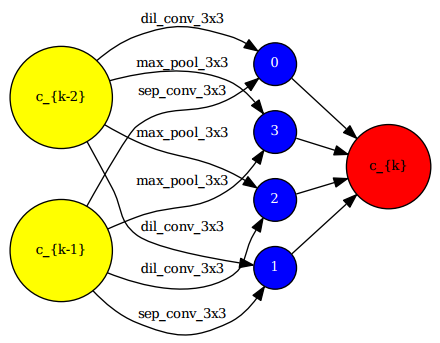}  
       \caption{Epoch 60}
       \label{fig:sub-second}
     \end{subfigure}
     \begin{subfigure}[b]{0.3\textwidth}
       \centering
       \includegraphics[width=\textwidth]{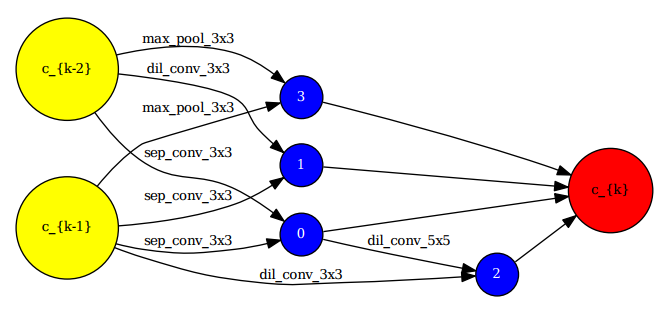}  
       \caption{Epoch 80}
       \label{fig:sub-second}
     \end{subfigure}
     \begin{subfigure}[b]{0.3\textwidth}
       \centering
       \includegraphics[width=\textwidth]{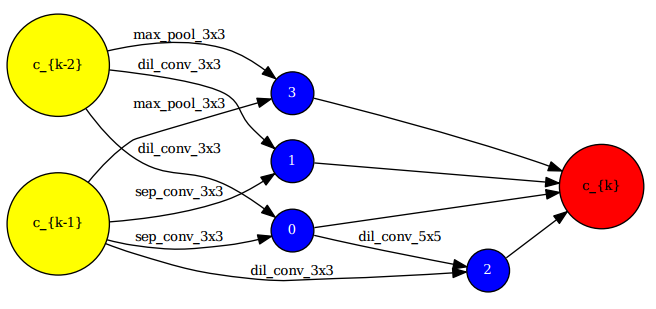}  
       \caption{Epoch 99}
       \label{fig:sub-second}
     \end{subfigure}
     \caption{Search progress of the normal cell in our method}
     \label{fig:search_progress}
   \end{figure*}
   \begin{table*}[h!]
   \centering
   \begin{tabular}{|c|c|c|c|c|c|c|}
   \hline
   Architecture       & (a)   & (b)  & (c)   & (d)   & (e)   & (f)   \\ \hline
   Parameter size (MB) & 2.89  & 2.89 & 3.07  & 2.89  & 3.27  & 3.09  \\ \hline
   Test acc. (\%)       & 96.87 & 97.15 & 97.14 & 97.15 & 97.22 & 97.21 \\ \hline
   \end{tabular}
   \caption{The parameter size and the test accuracy of the models in Figure~\ref{fig:search_progress}. }
   \label{table:search_progress}
   \end{table*}

\noindent \textbf{GPU hardware} 
 We conducted all the search experiments using a single NVIDIA Titan X (Pascal) GPU with 12GB memory for each run. In our method, a batch size of 64 was used due to the memory constraint of the GPU. By contrast, PDARTS used a P100 (Pascal) GPU with 16GB memory, which allows larger batch size for each run. For fair comparison of search cost, we run PDARTS on a Titan X GPU and report the performance with the constraint of 12GB memory capacity.

\subsection{Evaluation stage} \label{sec4.2}

\noindent \textbf{Architecture derivation}
The derivation algorithm in our method is adopted from DARTS. We consider the operation with top-1 architecture parameter as the optimal operation for each edge in one-shot network. Thus, the optimal cell is determined by pruning all operations other than the top-1 operation for each edge. The derived network is then retrained from scratch.

\noindent \textbf{Training settings}
The derived network is trained for 600 epochs. Other settings besides the number of training epochs are kept from the one-shot network training in the search stage.

\subsection{Transfer stage} \label{sec4.3}

\noindent \textbf{Large scale dataset} The network was based on the derived cell found by using a proxy dataset such as CIFAR-10 or CIFAR-100, and was evaluated on ImageNet dataset to test the transferability of our search algorithm.

   \begin{figure*}[htbp] 
     \centering
     \begin{subfigure}[b]{0.3\textwidth}
       \centering
       \includegraphics[width=\textwidth]{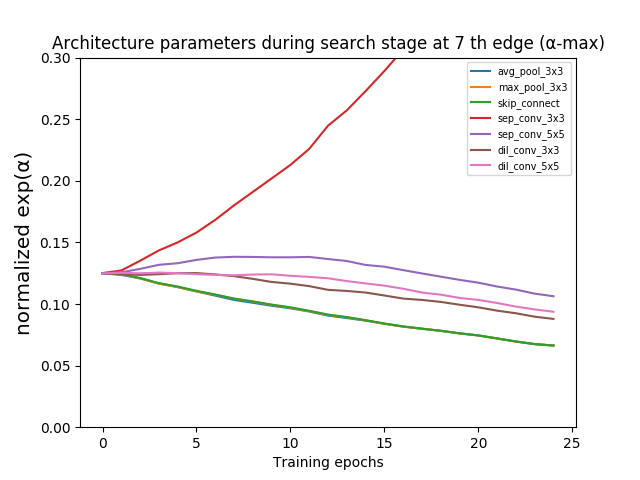}  
       \caption{$\alpha$-max, $7^{th}$ edge}
       \label{fig:sd1}
     \end{subfigure}
     \begin{subfigure}[b]{0.3\textwidth}
       \centering
       \includegraphics[width=\textwidth]{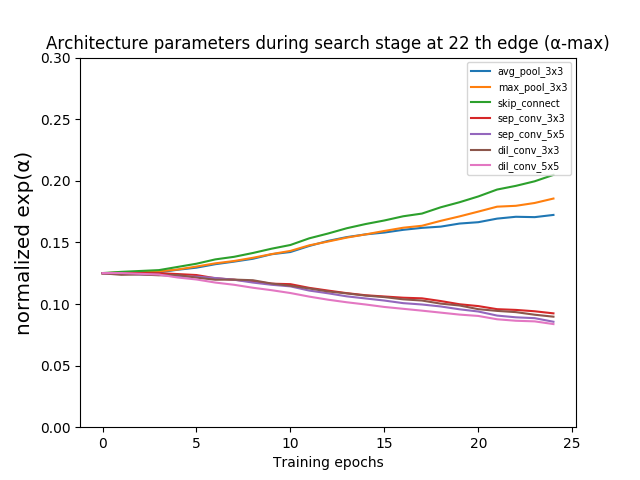}  
       \caption{$\alpha$-max, $22^{th}$ edge}
       \label{fig:sd2}
     \end{subfigure}
     \begin{subfigure}[b]{0.3\textwidth}
       \centering
       \includegraphics[width=\textwidth]{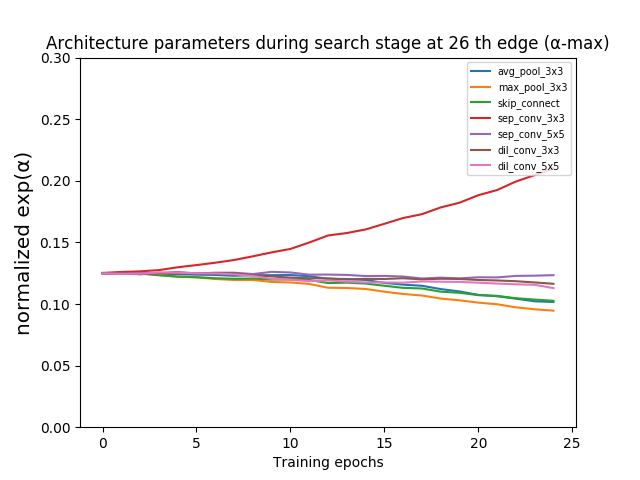}
       \caption{$\alpha$-max, $26^{th}$ edge}
       \label{fig:sd3}
     \end{subfigure}
     \begin{subfigure}[b]{0.3\textwidth}
       \centering
       \includegraphics[width=\textwidth]{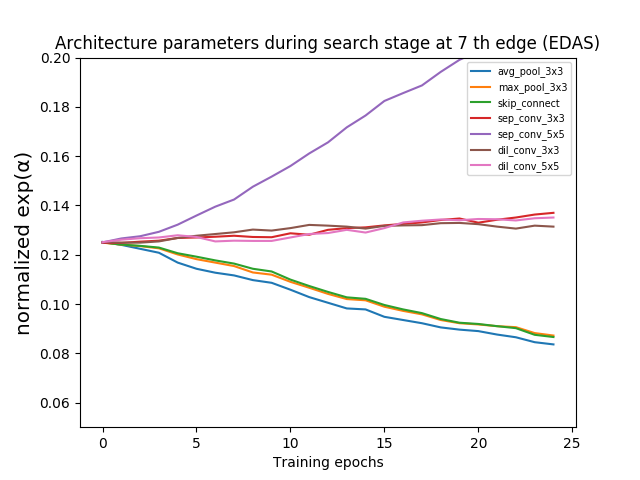}  
       \caption{EDAS, $7^{th}$ edge}
       \label{fig:sd4}
     \end{subfigure}
     \begin{subfigure}[b]{0.3\textwidth}
       \centering
       \includegraphics[width=\textwidth]{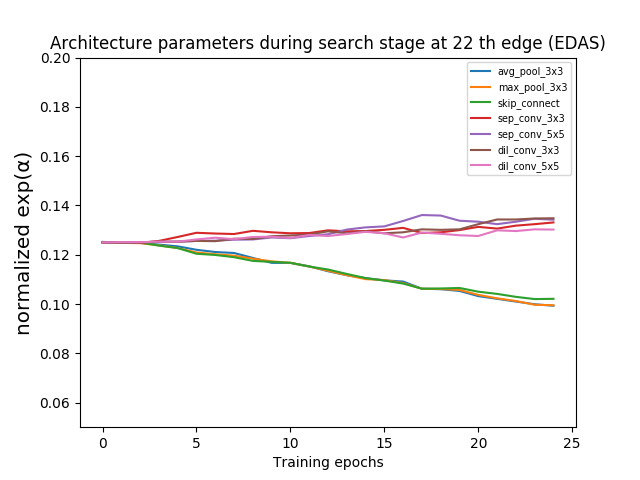}  
       \caption{EDAS, $22^{th}$ edge}
       \label{fig:sd5}
     \end{subfigure}
     \begin{subfigure}[b]{0.3\textwidth}
       \centering
       \includegraphics[width=\textwidth]{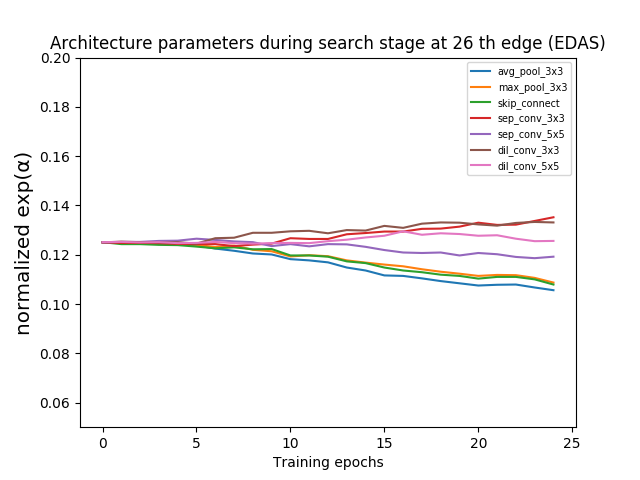}  
       \caption{EDAS, $26^{th}$ edge}
       \label{fig:sd6}
     \end{subfigure}
     \begin{subfigure}[b]{0.3\textwidth}
       \centering
       \includegraphics[width=\textwidth]{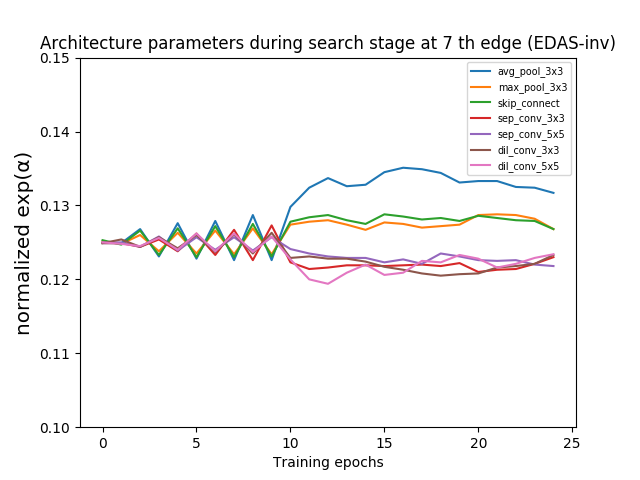}  
       \caption{EDAS-inv, $7^{th}$ edge}
       \label{fig:sd7}
     \end{subfigure}
     \begin{subfigure}[b]{0.3\textwidth}
       \centering
       \includegraphics[width=\textwidth]{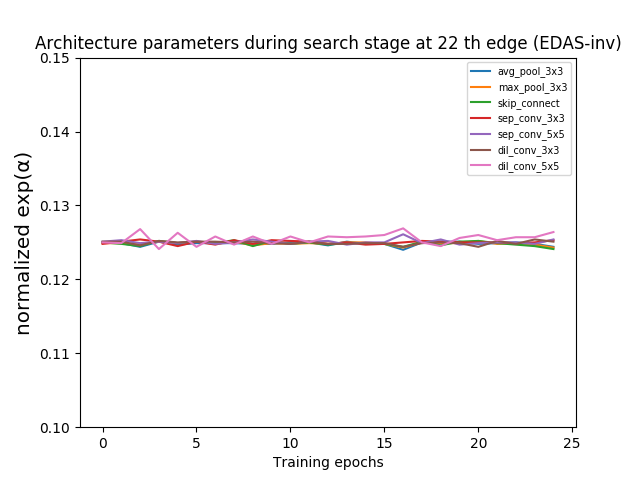}  
       \caption{EDAS-inv, $22^{th}$ edge}
       \label{fig:sd8}
     \end{subfigure}
     \begin{subfigure}[b]{0.3\textwidth}
       \centering
       \includegraphics[width=\textwidth]{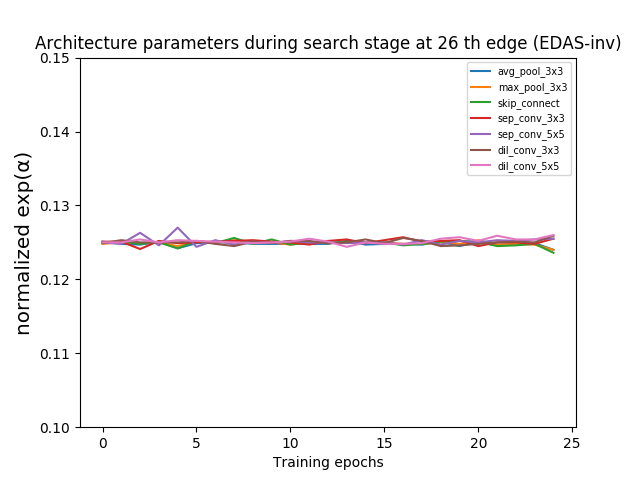}
       \caption{EDAS-inv, $26^{th}$ edge}
       \label{fig:sd9}
     \end{subfigure}
     \caption{Architecture parameter dynamics during search stage for $\alpha$-max method (top), EDAS (center), and EDAS-inv (bottom). Edges are indexed by numbering 1-14 to all the edges in normal cell and numbering 15-28 to all the edges in reduction cell in lexicographic order.}
     \label{fig:alpha}
   \end{figure*}

\begin{table*}[t]
\small
\begin{center}
\begin{tabular}{l|c|c|c|c}
    \toprule
    \multirow{2}{*}{\textbf{Architecture}} & \underline{\textbf{Test Err. (\%)}}  & \multirow{2}{*}{\textbf{Params (M)}} & \textbf{Search Cost} & \multirow{2}{*}{\textbf{Search Method}} \\
     & \textbf{C10 \quad C100} &  & \textbf{(GPU days)} &  \\
    \hline
    \midrule
    DenseNet-BC~\cite{huang2017densely}                 & \text{3.46 \quad 17.18} & 25.6 & - & manual  \\
    \midrule
    NASNet-A + cutout~\cite{zoph2018learning}           & \text{\enskip 2.65 \quad \quad  - \quad} & 3.3 & 1800 & RL  \\
    AmoebaNet-A + cutout~\cite{real2019regularized}     & \text{\enskip 3.34 \quad \quad  - \quad} & 3.2 & 3150 & evolution  \\
    AmoebaNet-B + cutout~\cite{real2019regularized}     & \text{\enskip 2.55 \quad \quad  - \quad}  & 2.8 & 3150 & evolution  \\
    Hierarchical Evo~\cite{liu2017hierarchical}         & \text{\enskip 3.75 \quad \quad  - \quad} & 15.7 & 300 & evolution  \\
    PNAS~\cite{liu2018progressive}                      & \text{\enskip 3.41 \quad \quad  - \quad} & 3.2 & 225 & SMBO  \\
    ENAS + cutout~\cite{pham2018efficient}                       & \text{\enskip 2.89 \quad \quad  - \quad} & 4.6 & 0.5 & RL  \\
    \midrule
    DARTS (first order) + cutout~\cite{liu2018darts}    & \text{3.00 \quad 17.76} & 3.3 & 1.5 & gradient-based  \\
    DARTS (second order) + cutout~\cite{liu2018darts}   & \text{2.76 \quad 17.54}  & 3.3 & 4.0 & gradient-based  \\
    SNAS + mild constraint + cutout~\cite{xie2018snas}   & \text{\enskip 2.98 \quad \quad  - \quad} & 2.9 & 1.5 & gradient-based  \\
    SNAS + moderate constraint + cutout~\cite{xie2018snas}   & \text{\enskip 2.85 \quad \quad  - \quad} & 2.8 & 1.5 & gradient-based  \\
    SNAS + aggressive constraint + cutout~\cite{xie2018snas}   & \text{\enskip 3.10 \quad \quad  - \quad} & 2.3 & 1.5 & gradient-based  \\
    ProxylessNAS + cutout~\cite{xie2018snas}   & \text{\enskip 2.08 \quad \quad  - \quad} & 5.7 & 4.0 & gradient-based  \\
    P-DARTS CIFAR10 + cutout~\cite{chen2019progressive}    & \text{2.50 \quad 16.55} & 3.4 & 0.3 & gradient-based  \\
    P-DARTS CIFAR100  + cutout~\cite{chen2017deeplab}    & \text{2.62 \quad 15.92} & 3.6 & 0.3 & gradient-based  \\
     P-DARTS CIFAR10-64 w/o architecture refinement $^\dag$ + cutout~\cite{chen2019progressive}         & \text{\enskip 6.17 \quad \quad  - \quad} & 1.37  & 0.32 & gradient-based  \\
    \midrule
    EDAS CIFAR10 + cutout                                     & \text{\enskip 2.92 \quad \quad  - \quad} & 3.44 & 0.125 & gradient-based  \\
    EDAS CIFAR10 warmup + cutout                                     & \text{\enskip 2.86 \quad \quad - \quad} & 3.15 & 0.125 & gradient-based  \\
    EDAS CIFAR100 + cutout                                     & \text{\quad \enskip -  \quad \enskip 17.61 \enskip} & 3.18 & 0.125 & gradient-based  \\
    EDAS CIFAR100 warmup + cutout                                     & \text{\quad \enskip -  \quad \enskip 17.16 \enskip} & 4.06 & 0.125 & gradient-based  \\
    \bottomrule
    
\end{tabular}
\end{center}
\caption{Comparison with state-of-the-art search algorithms on proxy datasets. $^{\dag}$ indicates P-DARTS method that excludes heuristic architecture refinement.}
\label{tab:sota}
\end{table*}

\section{Results} \label{sec5}

\subsection{Evaluation} \label{sec5.1}

In Table~\ref{tab:sota}, we compare our method with current state-of-the-art algorithms. Our method is heavily based on DARTS in that we train the relaxed edges and construct the one-shot network exactly as in DARTS. The evaluation result shows that our method outperforms DARTS (see DARTS (first order) and EDAS CIFAR10), even though the search cost is reduced to almost 1/12. On a single GPU, our method runs in only 3 hours, thus allowing fast discovery of architectures needed for a new task. Compared to other efficient methods such as ENAS and SNAS, EDAS shows better or comparable results with much a smaller search cost. Also, compared to P-DARTS, which is another state-of-the-art method in terms of search cost and test accuracy, our method still shows an advantage in efficiency with small performance gap.

We also report evaluation results with architectures discovered with CIFAR-100 dataset. P-DARTS was the first work that published search result on CIFAR-100, because previous methods do not show improved performance on CIFAR-100 dataset. Our result shows that EDAS can also be used for search on CIFAR-100 with comparable result to DARTS.

\subsection{Search behavior} \label{sec5.2}

In Figure~\ref{fig:search_progress}, we report how the discovered cell changes qualitatively. We run our search algorithm for 100 epochs to observe search behaviour precisely. As a result, the network depth decreases in earlier epochs. (See Figure~\ref{fig:search_progress} (b)-(d).) We attribute this phenomena to easier optimization from shallower network. However, network depth increases once in the middle of training to obtain better performance. (See Figure~\ref{fig:search_progress} (e)-(f).) The configuration of operations changes suddenly in the first few epochs of training, but it gradually changes in the remaining epochs. We can see that the derived architectures in Figure~\ref{fig:search_progress} (b)-(f) are similar.

We also conduct a quantitative analysis on search progress in terms of parameter size and test accuracy of the models obtained from the cells in Figure~\ref{fig:search_progress}. As described in Table~\ref{table:search_progress}, the test accuracy of the discovered network increases as the search progresses in our method, thus demonstrating the search ability of our algorithm. However, there is no distinct pattern in parameter size change of the derived architecture during \textit{search stage}.

\subsection{Architecture parameter dynamics} \label{sec5.3}
In this section, we describe how architecture parameter changes during \textit{search stage} for several possible methods in Figure~\ref{fig:alpha}. The methods include \textbf{1)}$\mathbf{\alpha}$-\textbf{max}, \textbf{EDAS}, and \textbf{EDAS-inv}. Basically, all the methods learn architecture parameters by sampling a single edge. $\mathbf{\alpha}$-\textbf{max} trains operation weights by sampling the operation with the maximum architecture parameter ($\alpha$) for each edge. \textbf{2) EDAS} is our first method that samples the operation with the maximum $\alpha$ for each edge except the sampled edge that is relaxed. \textbf{3) EDAS-inv} is our improved method that is the same with EDAS but only inversing the sign of all the architecture parameters for each iteration of the first few epochs to give a fair learning chance to all the candidate operations. 

We randomly sampled three edges among all the edges within both normal cells and reduction cells and showed how architecture parameter changes during search stage for each method mentioned above in Figure~\ref{fig:alpha}. We observed that the ranking of architecture parameter values does not change during search stage and the architecture is determined in the first few epochs for the case of $\mathbf{\alpha}$-\textbf{max}. In contrast, the ranking of architecture parameter values changes several times during search stage and it implies that \textbf{EDAS} explores the search space better than $\mathbf{\alpha}$-\textbf{max}. For the case of \textbf{EDAS-inv}, the ranking of architecture parameter values changes dynamically during the first few epochs where the sign of each architecture parameter is inversed to give less experience to the operation with higher $\alpha$. Based on fair experiences among candidate operations, $\alpha$ ranking keeps changing steadily for the remaining epochs until the end of search stage. Thus, we attribute the performance elevation of \textbf{EDAS-inv} from \textbf{EDAS} to better exploration power of \textbf{EDAS-inv}.

\section{Conclusion} \label{sec6}

In this paper, we propose an efficient differentiable one-shot architecture search algorithm by sampling a single edge in the cell for update. This simple modification, combined with proper regularization, improves performance over DARTS with search cost of only 0.125 GPU days. Through extensive experiments and detailed analysis, we show that our method not only outperform other methods, but also is capable of training meaningful weights in the search stage. Our work can also be widely applied to other methods based on differentiable architecture search.

{\small
\bibliographystyle{ieee_fullname}
\bibliography{egbib}
}

\end{document}